%% file: main.tex
\documentclass{article}
\usepackage{iclr2027_conference,times}
\usepackage{amsmath,amssymb,graphicx,booktabs,tabularx,xcolor}
\usepackage{wrapfig,colortbl}
\usepackage{placeins,needspace}
\usepackage{microtype}
\usepackage{hyperref,url}
\usepackage{trimclip}
\usepackage{tikz,fontawesome5}
\definecolor{evokeborder}{HTML}{B5CAD6}
\newcommand{\evokebadge}[2]{\tikz[baseline=(badge.base)]\node[draw=evokeborder,line width=0.4pt,rounded corners=8pt,inner xsep=8pt,inner ysep=3pt,font=\sffamily\bfseries\small](badge){#1\hspace{4pt}#2};}
\newcommand{\evokelink}[2]{{\hypersetup{pdfborder={0 0 0}}\href{#1}{#2}}}
\newcommand{\evokeresources}{\evokelink{https://gnonymous.github.io/EVOKE}{\evokebadge{\textcolor{evoketitle}{\faGlobe}}{Project Page}}\hspace{12pt}\evokelink{https://github.com/Gnonymous/EVOKE}{\evokebadge{\faGithub}{Code}}\hspace{12pt}\evokelink{https://huggingface.co/Gnonymous/EVOKE}{\evokebadge{\raisebox{-1.5pt}{\includegraphics[height=10pt]{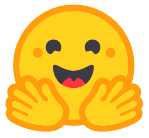}}}{Model}}}
\definecolor{evoketitle}{HTML}{007BFF}
\newsavebox{\evokewbox}
\DeclareRobustCommand{\evokehalfW}{\begingroup
\sbox{\evokewbox}{W}%
\makebox[\wd\evokewbox][l]{\rlap{\usebox{\evokewbox}}\clipbox{0pt 0pt {0.5\width} 0pt}{\textcolor{evoketitle}{W}}}%
\endgroup}
\title{\centering\raisebox{-0.12em}[0pt][0pt]{\includegraphics[height=1.05em,trim=320 190 330 245,clip]{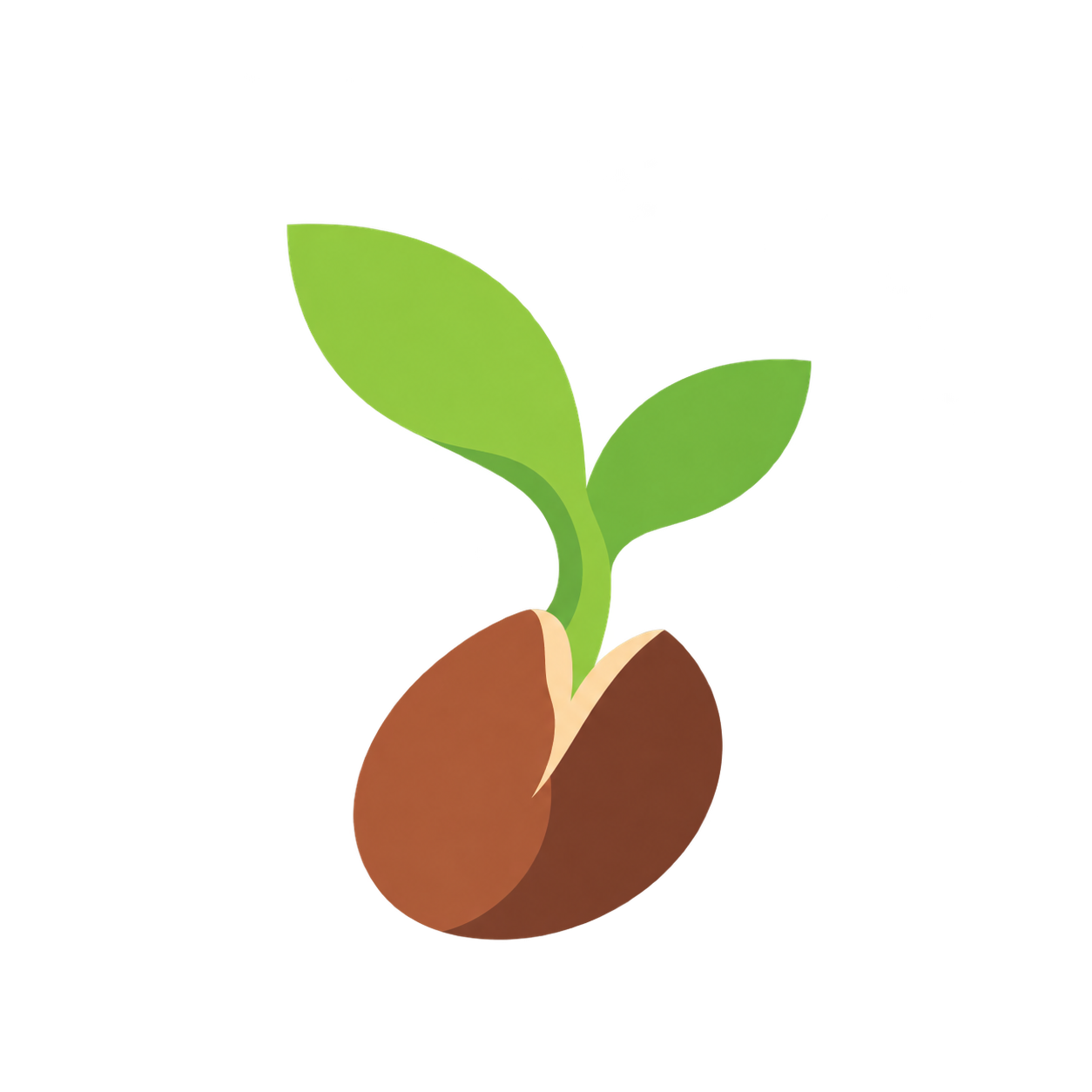}}\hspace{0.25em}\textcolor{evoketitle}{EVOKE}: \textcolor{evoketitle}{E}liciting \evokehalfW\textcolor{evoketitle}{o}rld \textcolor{evoketitle}{K}nowledg\textcolor{evoketitle}{e} in Agents for Transferable Decision-Making}
\author{\parbox{\dimexpr\textwidth-2\tabcolsep\relax}{\centering\normalfont
{\normalsize\bfseries Yuhan Guo$^{1,2}$, Jinming Liu$^{1,2}$, Liang Xu$^{3}$, Ziqiang Li$^{1,2}$, Jianguo Huang$^{1,2}$,\\
Zhicheng Wang$^{2,4}$, Hu Zhu$^{2,4}$, Qiuyu Chen$^{1,2}$, Yuntao Wei$^{2,4}$, Xin Jin$^{2}$, Wenjun Zeng$^{2}$}\\[2pt]
{\normalsize $^{1}$Shanghai Jiaotong University\quad
$^{2}$Eastern Institute of Technology, Ningbo\\
$^{3}$Zhongguancun Academy, Beijing, China\quad
$^{4}$Hong Kong Polytechnic University}
}}
\iclrfinalcopy
\usepackage{etoolbox}
\makeatletter
\patchcmd{\@maketitle}{\@title\par}{\@title\par\vskip 8pt}{}{}
\patchcmd{\@maketitle}{\vskip 0.3in minus 0.1in}{\par\kern12pt\nointerlineskip\vbox to\dimexpr 0.3in-8pt\relax{\vss\hbox to\textwidth{\hfil\evokeresources\hfil}\vss}}{}{}
\makeatother
\newcommand{\evoke}{\textsc{Evoke}}
\definecolor{evokebest}{HTML}{E8E3F1}
\definecolor{evokesecond}{HTML}{E6EFF6}
\DeclareRobustCommand{\bestresult}{\begingroup\setlength{\fboxsep}{1pt}\colorbox[HTML]{E8E3F1}{\textbf{best}}\endgroup}
\DeclareRobustCommand{\secondresult}[1][E6EFF6]{\begingroup\setlength{\fboxsep}{1pt}\colorbox[HTML]{#1}{\underline{second-best}}\endgroup}

\begin{document}
\vspace*{-24pt}
\maketitle
\fancyhead{}
\begin{abstract}

\input{abstract}
\end{abstract}
\input{sections/1_introduction_version2}
\input{sections/2_related_work}
\input{sections/3_method}
\input{sections/4_experiments}

\vspace{3mm}
\input{sections/5_analysis}
\vspace{3mm}
\input{sections/6_conclusion}
\label{maintextend}
\clearpage
\input{statements}
\bibliography{references}
\bibliographystyle{iclr2027_conference}
\clearpage
\appendix
\input{appendix}
\end{document}

%% file: abstract.tex
Large language models (LLMs) are increasingly deployed as agents for multi-step decision-making, yet transfer poorly to unseen environments. World-model methods address this by training agents to predict future observations, at the cost of additional training and errors that compound when predictions are used for planning. However, for LLM agents operating in digital environments, much of this world knowledge is already internalized during pretraining, which shifts the problem from acquiring it to eliciting it. We argue that typical post-training provides little pressure for such elicitation, since supervision under a single goal at each visited state inadvertently drives policies to rely on superficial contextual habits. We introduce \textbf{\evoke{}}, a post-training method that supplies this pressure through goal diversity at fixed states. Motivated by theory showing that an agent competent across diverse goals must encode a world model recoverable from its action preferences, \evoke{} holds the environment state and interaction history fixed and ranks the same candidate actions under alternative goals, forcing action preferences to change, so that a policy relying on contextual habits or single-goal correlations cannot order them correctly. This implicitly elicits the policy's pretrained world knowledge to inform decisions. We evaluate \evoke{} across diverse tasks in three backbones, demonstrating improved task performance, unseen environment generalization, and data efficiency. We further conduct controlled analyses to better understand what drives these gains. These findings offer a new perspective on eliciting internalized world knowledge for transferable action through direct decision supervision.

%% file: sections/1_introduction_version2.tex
\section{Introduction}
\label{sec:intro}

Large language models (LLMs) are increasingly post-trained as agents to make multi-step decisions in interactive environments \citep{yao2023react,liu2026gem,liu2025survey}. These agents often perform well where they were trained, yet their performance drops considerably in environments they have not seen \citep{zhang2026generalizability}. What they learn stays tied to the tasks and environments encountered during training and rarely transfers to new ones.

\begin{wrapfigure}[31]{r}{0.49\textwidth}
\vspace{-2pt}
\setlength{\abovecaptionskip}{4pt}
\setlength{\belowcaptionskip}{0pt}
\centering
\includegraphics[width=\linewidth]{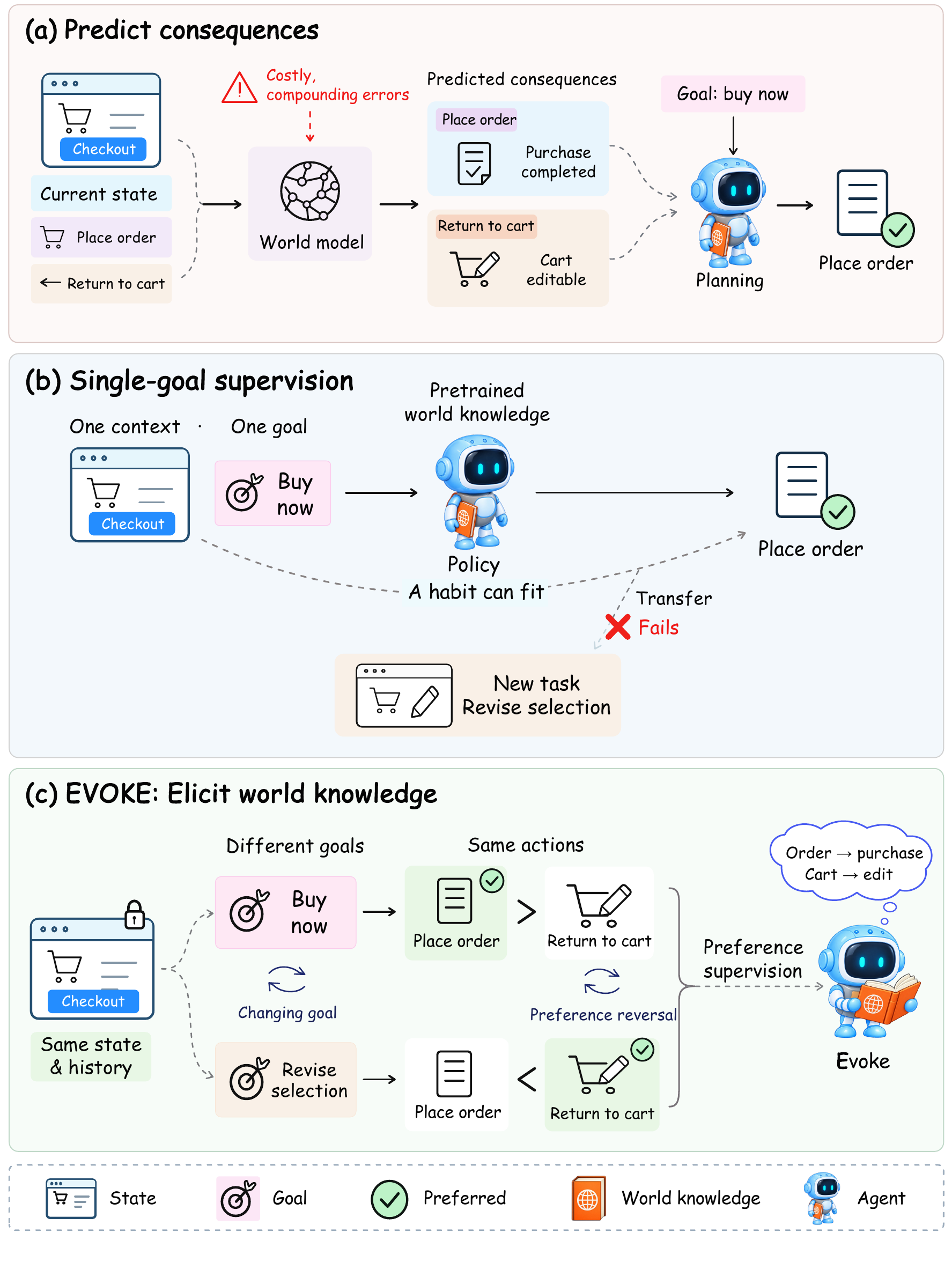}
\small
\caption{\textbf{From prediction to preference.} (a) World-model approaches predict action consequences before choosing an action. (b) Under single-goal supervision, a policy can fit habits that fail to transfer. (c) \evoke{} instead supervises action preferences at a fixed state and history under different goals.}
\label{fig:concept}
\end{wrapfigure}

One way to improve transferability is to let the agent anticipate the consequences of an action before choosing it. The effect of an action often follows rules that hold across environments (e.g., placing an order ends a purchase on any shopping site), so decisions that account for these effects are more likely to carry over. World-model methods pursue this idea by training agents to predict the consequences of their actions, in the form of future observations in Figure~\ref{fig:concept}a \citep{zhang2026iwm,lu2026,wang2026envrl,liu2026itp}. However, learning these consequences through prediction comes with two costs. (1) It adds a prediction objective that makes training more expensive, and much of what the agent is asked to predict is irrelevant to the decision at hand \citep{grimm2020}. (2) When predicted futures are used for planning, their errors compound over successive steps \citep{zhou2025walle2,liu2026itp}.

Yet do LLM agents need to learn these consequences through prediction? Such learning is important in physical environments, where dynamics such as contact and motion are hard to capture without grounding \citep{assran2025}. LLM agents, however, mostly act in digital environments such as websites and search engines, where the effects of actions fall largely within the world knowledge that LLMs acquire through pretraining \citep{brown2020fewshot}. Indeed, recent work finds that pretrained
\newpage
\noindent
 models, given only a few examples, can already predict how text-based environments respond to many actions \citep{li2026wordworld}. This suggests that LLM agents may not need to rely entirely on a world model to predict, since they already know much of the world knowledge from pretraining.

\textit{This shifts the problem from acquiring this knowledge to eliciting it, ensuring the model actually leverages what it already knows to make decisions.} However, typical training paradigms inadvertently limit this elicitation by driving the policy toward surface-level behavior matching. When a visited state is tied to a training goal, the policy can easily fit the training signal just by memorizing the habitual next action for that context (Figure~\ref{fig:concept}b). While these superficial habits work within the training distribution, they quickly break down in new environments, reflecting how single-goal supervision provides little pressure to use pretrained knowledge at decision time.

We introduce \evoke{}, a post-training method that supplies this pressure through goal diversity at fixed states. \citet{richens2025} motivate this choice, showing that an agent competent across a sufficiently rich set of multi-step goals must contain a world model recoverable from its goal-conditioned behavior. Targeting this competence, \evoke{} holds the environment state and interaction history fixed while introducing alternative goals (Figure~\ref{fig:concept}c), thereby evaluating the exact same set of candidate actions against different objectives. Crucially, this isolates the goal's effect and often forces action preferences to flip. Whenever the preferred actions differ across these alternative goals, a policy that relies on contextual habits from standard supervision or on correlations learned under a single fixed goal cannot rank them correctly. This encourages the policy to elicit its pretrained world knowledge of action consequences into its decision-making.

In practice, \evoke{} collects states from the policy's own rollouts, poses alternative goals at these states, and trains the policy to rank candidate actions under each goal. This process is repeated over rounds in the manner of DAgger \citep{ross2011}. Across household tasks, web navigation, and search-based QA, \evoke{} demonstrates transferability, yielding improved performance, unseen environment generalization and data efficiency. To isolate the mechanisms driving these gains, we conduct extensive controlled analyses. Ultimately, we hope these findings offer a new perspective on eliciting internalized world knowledge for transferable action. Taken together, our work makes the following contributions:
\begin{enumerate}
\setlength{\itemsep}{2pt}
    \item We revisit how LLM agents can benefit from world knowledge, framing the problem as eliciting knowledge already present in pretrained models rather than acquiring it through additional prediction objectives.
    \item We introduce \evoke{}, which explores goal diversity at fixed states as a source of decision-level supervision. By ranking the same candidate actions under different goals, it encourages the policy to bring the knowledge of action consequences acquired in pretraining into its decisions.
    \item We evaluate \evoke{} on text-based household tasks, web navigation, and search-based QA across three backbones, observing gains in task performance, generalization to unseen environments, and data efficiency. We further conduct controlled analyses to better understand what drives these gains.
\end{enumerate}

%% file: sections/2_related_work.tex
\section{Related Work}
\label{sec:related}
\paragraph{World models for LLM agents.}
A line of work equips LLM agents with knowledge of action consequences by learning to predict them. Some methods build an explicit model for lookahead: WALL-E combines pretrained knowledge with symbolic rules, ITP learns textual dynamics for adaptive lookahead, and MemWM adds memory \citep{zhou2025walle2,liu2026itp,wang2026memwm}. Others train the policy with an auxiliary prediction signal: IWM learns next-state prediction before imitation learning, PaW jointly optimizes observation prediction and policy learning, EnvRL adds forward and inverse dynamics, and Role-Agent and RWML derive rewards from predicted--observed state agreement \citep{zhang2026iwm,lu2026,wang2026envrl,wang2026roleagent,yu2026rwml}. In embodied settings, V-JEPA~2's action-conditioned extension predicts latent representations for robotic planning \citep{assran2025}. In a complementary knowledge-based approach, WKM trains a separate world knowledge model to guide planning with task-level and state-level knowledge \citep{qiao2025wkm}. From Word to World finds that pretrained LLMs, given a few demonstrations, already predict next states well in structured text-based environments \citep{li2026wordworld}. \evoke{} builds on this observation and elicits the consequence knowledge already present in the policy, without a prediction objective or a separate model.\looseness=-1

\paragraph{Policy optimization for LLM agents.}
LLM agents are commonly improved from their own interaction. Reinforcement learning optimizes the policy with task rewards, using group-relative estimates at the trajectory or step level \citep{shao2024grpo,feng2025gigpo}, and distillation-based methods add denser guidance from teacher signals \citep{lu2026sdar}. ETO learns from preferences between successful and failed trajectories \citep{song2024eto}, and DAgger aggregates labels at learner-visited states \citep{ross2011}. These methods mainly differ in how the training signal is computed. \evoke{} focuses on a complementary aspect: the goals under which each visited state is supervised. When a state is supervised under a single goal, the policy can fit the signal through contextual habits (Section~\ref{sec:intro}); \evoke{} supplies goal diversity at fixed states to add the pressure that single-goal supervision leaves out.\looseness=-1

\paragraph{Learning across goals.}
Goal-conditioned reinforcement learning shares experience across goals. Universal value functions generalize over goals \citep{schaul2015}, successor features separate environment dynamics from task-specific rewards to enable transfer \citep{barreto2017}, and Hindsight Experience Replay relabels trajectories with achieved goals so that failures still provide learning signals \citep{andrychowicz2017}. Hindsight Supervised Learning brings relabeling to language-agent trajectories by constructing demonstrations for achieved goals \citep{li2026hsl}. \citet{richens2025} show that an agent competent across a sufficiently rich set of goals must contain a world model recoverable from its goal-conditioned behavior. \evoke{} applies goal diversity to a single decision: it holds the state and history fixed, assesses the same candidate actions under alternative goals, and trains on the resulting action preferences rather than on relabeled trajectories alone.\looseness=-1

%% file: sections/3_method.tex
\section{EVOKE}
\label{sec:method}
\begin{figure}[!htb]
\centering
\includegraphics[width=\linewidth]{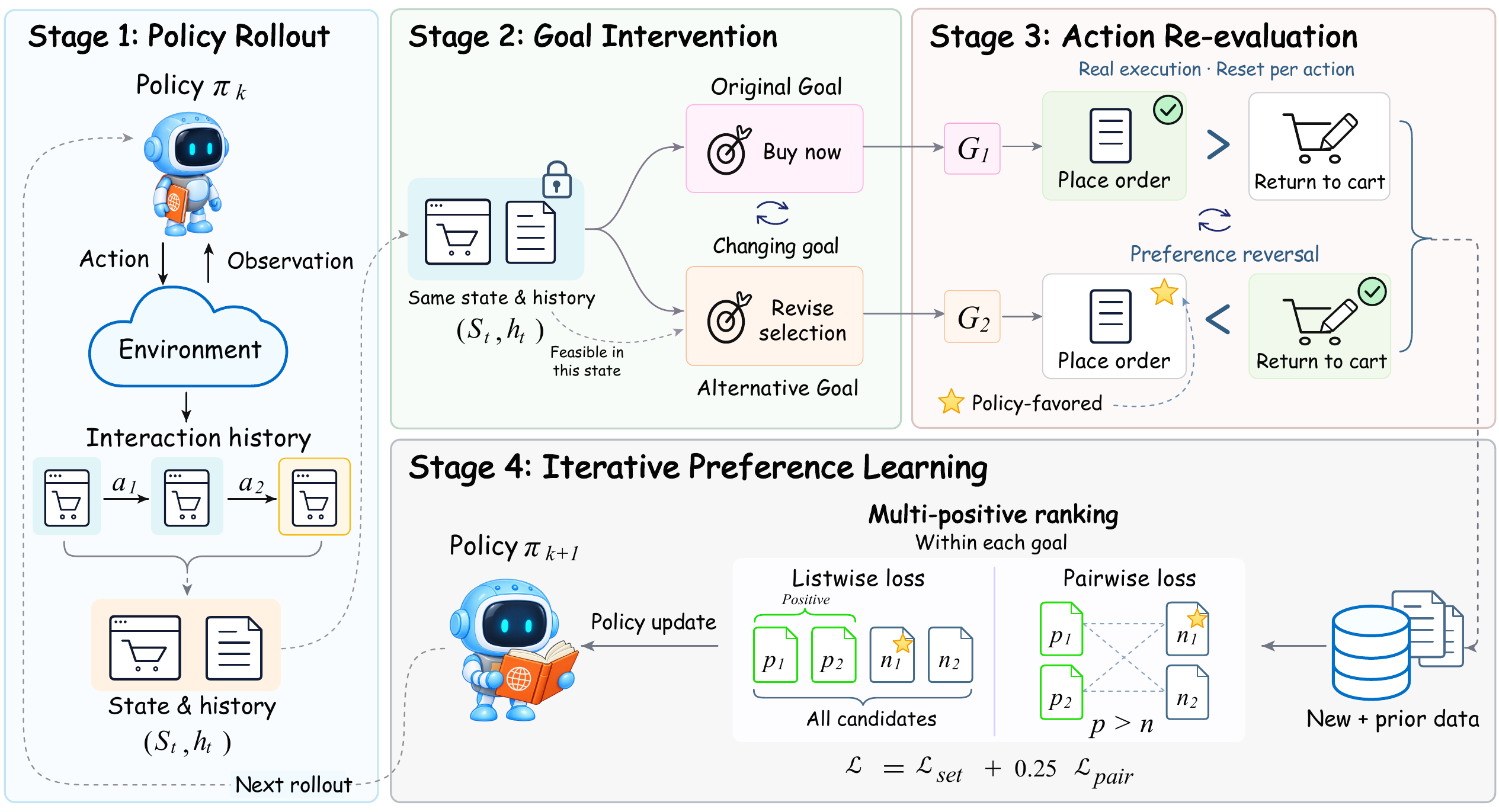}
\caption{\textbf{The \evoke{} training loop.} At collected states, actions are executed and assessed under alternative goals with the state, history, and available actions fixed. The policy learns by contrastive ranking on aggregated preferences; actions can switch between positive and competing across goals.}
\label{fig:pipeline}
\end{figure}

\evoke{} trains a language-model policy $\pi_\theta$ that selects the next action from the input $x_g=(g,h,\mathcal A(s))$, consisting of a goal $g$, the visible interaction history $h$, and the actions $\mathcal A(s)$ available in the current state $s$. Training repeats a loop of state collection, goal intervention, action assessment, and contrastive ranking (Figure~\ref{fig:pipeline}); at deployment, \evoke{} is a standard policy.

\subsection{Goal Interventions at Fixed States}
\label{sec:goal-intervention}
Changing the goal does not change how the environment responds to an action, but it changes which responses matter. We write this as
\begin{equation}
T_g(s'\mid s,a)=T(s'\mid s,a),\qquad
J_g(s,h,a)=\sum_{s'}T(s'\mid s,a)\,u_g(h,s,a,s'),
\label{eq:decomposition}
\end{equation}
where $T$ is the transition shared by all goals and $u_g$ scores how useful a consequence is for goal $g$. Equation~\ref{eq:decomposition} separates what an action does, which is common across goals, from what it is worth, which is specific to each goal.

Single-goal supervision leaves this structure unused: when each visited context $(s,h)$ is labeled under its original goal only, the policy can fit the labels by associating familiar contexts with habitual next actions. A goal intervention holds $s$, $h$, and $\mathcal A(s)$ fixed and replaces $g$ with an alternative goal $g'$. In Figure~\ref{fig:pipeline}, the agent is at a checkout page: under the goal of buying now, it should place the order, whereas under the goal of revising the selection, it should return to the cart. Whenever the preferred actions differ, no mapping from the context alone ranks the candidates correctly under both goals. Ranking correctly then requires conditioning on the goal, most directly by relating what each action does, which is shared across goals, to what the current goal requires. Goal interventions thus pressure the policy to use its knowledge of action consequences, rather than leaving it merely available.

\subsection{Constructing Goal-Conditioned Action Preferences}
\label{sec:construction}
\paragraph{Collecting decision states.}
We initialize the policy, denoted $\pi_{\mathrm{boot}}$, by supervised fine-tuning on successful demonstrations decomposed into per-step input--action pairs. The policy then interacts with training environments, and we record the goal and action--observation history at each step, so that the same state can be revisited. We sample decision states where the policy makes progress as well as where it takes detours, repeats actions, or heads toward failure, so supervision concentrates on situations the policy actually encounters.\looseness=-1

\paragraph{Goal interventions.}
For each collected state, we keep the original goal and add alternative goals achievable from the same state. An LLM annotator proposes candidates; each is instantiated in the environment together with its success condition and retained if it is compatible with the scene and not yet satisfied. The state, visible history, and available actions stay identical across goals.

\paragraph{Grounded action assessment.}
Candidate actions are drawn from the shared action set $\mathcal A(s)$ (Appendix~\ref{app:labeling}), and each is executed from the same state, with a short follow-up interaction when one step does not reveal its effect. Given the goal, the history, and the executed outcomes, the annotator labels actions that advance the goal as positives $P$ and the others as competitors $N$, leaving actions with insufficient evidence unlabeled. Because the annotator judges observed outcomes rather than predicting them, the labels reflect how the environment actually responds.

\paragraph{Policy-favored hard negatives.}
We fill each competitor set first with actions that the policy favors but the annotator judges worse: the policy's greedy action when it is not a positive, followed by other high-probability non-positive actions, and then actions of other types for coverage. These are the mistakes the policy is most likely to make, and thus the most informative negatives \citep{robinson2021}. For a goal of putting a clean cup on the table while holding a dirty cup, going to the sink is a positive, whereas placing the dirty cup directly on the table, which the policy readily proposes, is a hard negative.

\subsection{Learning through Contrastive Ranking}
\label{sec:ranking}
We score an action $a$ by the mean token log-probability of its completion $c(a)$,
\begin{equation}
z_\theta(a\mid x_g)=\frac{1}{|c(a)|}\sum_{t=1}^{|c(a)|}
\log \pi_\theta\bigl(c_t(a)\mid x_g,c_{<t}(a)\bigr),
\end{equation}
and write $r_a=z_\theta(a\mid x_g)/\tau$ with temperature $\tau$. For a group with positives $P$ and competitors $N$, the loss is $\mathcal L=\mathcal L_{\mathrm{set}}+\lambda\mathcal L_{\mathrm{pair}}$, where $\lambda$ weights the pairwise term and
\begin{equation}
\mathcal L_{\mathrm{set}}=-\log \frac{\sum_{p\in P} e^{r_p}}{\sum_{a\in P\cup N} e^{r_a}},\qquad
\mathcal L_{\mathrm{pair}}=-\frac{1}{|P||N|}\sum_{p\in P}\sum_{n\in N}\log\sigma(r_p-r_n),
\label{eq:loss}
\end{equation}
and $\sigma$ is the logistic function. The listwise term $\mathcal L_{\mathrm{set}}$ raises the mass of the positive set as a whole, so several actions can be correct in one context; the pairwise term separates every positive from every competitor. Supervised fine-tuning on positives raises preferred actions but does not explicitly push down the policy's own tempting mistakes, whereas both ranking terms push these hard negatives down. The loss is computed within each goal; the relation across goals comes from contexts that share a state but differ in their positives. Executed outcomes and annotator judgments build labels only and never enter the policy input. We update LoRA adapters \citep{hu2021} with $\tau=1$ and $\lambda=0.25$; further details are in Appendix~\ref{app:labeling}.\looseness=-1

\subsection{Iterative Aggregation}
\label{sec:iteration}
An updated policy reaches new states and makes new mistakes. Each round therefore repeats state collection, goal intervention, action assessment, and hard-negative selection with the current policy, and trains on the aggregate of all rounds, following dataset aggregation in imitation learning \citep{ross2011} (per-round data statistics in Appendix Table~\ref{tab:supervision}). This keeps the hard negatives aligned with the errors of the current policy. At deployment, the policy acts from the goal, history, and available actions alone, with no world-model module, no inference-time planning, and no annotator.

%% file: sections/4_experiments.tex
\section{Experiments}
\label{sec:experiments}
We evaluate whether goal-conditioned decision supervision improves complete task execution in unseen environments, then examine which training choices account for the gains. The main comparison covers three model backbones. Section~\ref{sec:analysis} separately studies supervision design and training budget using a shared 3B initialization.

\subsection{Setup and Evaluation Protocol}
\paragraph{Benchmarks and metrics.}
\label{sec:benchmark-metrics}
All scores are percentages, and EVOKE results are averaged over training seeds. \textbf{ALFWorld} reports success on Pick, Look, Clean, Heat, Cool, and Pick2 \citep{shridhar2021}; Avg is their unweighted mean. EVOKE uses 140 Seen games in the main table and 134 Unseen games separately, with frozen parameters and a 50-action limit.
\textbf{Search-based QA} reports answer accuracy on NQ \citep{kwiatkowski-etal-2019-natural}, TriviaQA \citep{joshi-etal-2017-triviaqa}, PopQA \citep{mallen-etal-2023-trust}, HotpotQA \citep{yang-etal-2018-hotpotqa}, 2WikiMultiHopQA \citep{ho-etal-2020-constructing}, MuSiQue \citep{trivedi-etal-2022-musique}, and Bamboogle \citep{press-etal-2023-measuring}; Avg equally weights these seven datasets.
\textbf{WebShop} \citep{yao2022webshop} reports mean task score (Score) and task success rate (Succ.).

\paragraph{Baselines.} We compare with the three groups in Tables~\ref{tab:main-aw}--\ref{tab:evoke-unseen}: prompting (Vanilla, ReAct~\citep{yao2023react}, Skill-Prompt); policy optimization and distillation (GRPO~\citep{shao2024grpo}, Skill-GRPO, OPSD~\citep{zhao2026opsd}, GRPO+OPSD, Skill-SD~\citep{wang2026skillsd}, RLSD~\citep{yang2026rlsd}, SDAR~\citep{lu2026sdar}, OPID~\citep{yang2026opid}, PCSD~\citep{lv2026pcsd}, GRSD~\citep{zheng2026grsd}, AHEAD~\citep{jin2026ahead}); and world-model and dynamics methods (IWM~\citep{zhang2026iwm}, PaW~\citep{lu2026}, EnvRL~\citep{wang2026envrl}, ITP~\citep{liu2026itp}, MemWM~\citep{wang2026memwm}).

\input{tables/main_results_improved}

\subsection{Main Results}
Tables~\ref{tab:main-aw} and~\ref{tab:main-qa-v2} compare methods on Qwen2.5-3B/7B-Instruct \citep{qwen2025qwen25technicalreport} and Qwen3-1.7B \citep{yang2025qwen3technicalreport} backbones. \evoke{} achieves the best average on every benchmark and backbone, including all world-model and dynamics methods on 7B, and remains the best on unseen ALFWorld games (Table~\ref{tab:evoke-unseen}). The margins are largest on ALFWorld and WebShop, which require multi-step interaction, and on the smallest backbone, Qwen3-1.7B. On search-based QA, where each question involves only a few retrieval decisions and the baselines are closer, \evoke{} still leads the strongest baseline on each backbone. Section~\ref{sec:analysis} analyzes where the gains come from.\looseness=-1

\vspace{-6mm}
\input{tables/evoke_alfworld_unseen}

%% file: tables/main_results_improved.tex
\definecolor{mainbest}{HTML}{E8E3F1}
\definecolor{mainblue}{HTML}{E6EFF6}
\definecolor{maingreen}{HTML}{E8F3ED}
\definecolor{mainyellow}{HTML}{F8F2D7}
\definecolor{mainbackbone}{HTML}{F5F7FA}
\providecommand{\improvedbest}[1]{\cellcolor{mainbest}\textbf{#1}}
\providecommand{\improvedsecondblue}[1]{\cellcolor{mainblue}\underline{#1}}
\providecommand{\improvedsecondgreen}[1]{\cellcolor{mainblue}\underline{#1}}
\providecommand{\improvedsecondyellow}[1]{\cellcolor{mainblue}\underline{#1}}
\newlength{\improvedAWrowextra}
\newlength{\improvedQArowextra}
\newlength{\improvedHeightDelta}
\newsavebox{\improvedAWmeasure}
\newsavebox{\improvedQAmeasure}
\begin{table}[!t]
\setlength{\parskip}{0pt}\setlength{\abovecaptionskip}{3pt}\setlength{\belowcaptionskip}{3pt}
\setlength{\improvedAWrowextra}{0pt}\setlength{\improvedQArowextra}{0pt}
\sbox{\improvedAWmeasure}{\def\improvedMeasurePrefix{}\begin{minipage}{0.515\linewidth}\input{tables/results_alfworld_webshop}\end{minipage}}
\sbox{\improvedQAmeasure}{\def\improvedMeasurePrefix{}\begin{minipage}{0.465\linewidth}\input{tables/results_search_qa}\end{minipage}}
\setlength{\improvedHeightDelta}{\dimexpr\ht\improvedAWmeasure+\dp\improvedAWmeasure-\ht\improvedQAmeasure-\dp\improvedQAmeasure\relax}
\ifdim\improvedHeightDelta>0pt
  \setlength{\improvedQArowextra}{\improvedHeightDelta}\divide\improvedQArowextra by 24
\else
  \setlength{\improvedAWrowextra}{-\improvedHeightDelta}\divide\improvedAWrowextra by 27
\fi
\begin{minipage}[t]{0.515\linewidth}
\caption{ALFWorld and WebShop (\%).}
\label{tab:main-aw}
\input{tables/results_alfworld_webshop}
\end{minipage}\hfill
\begin{minipage}[t]{0.465\linewidth}
\caption{Search-based QA accuracy (\%).}
\label{tab:main-qa-v2}
\input{tables/results_search_qa}
\end{minipage}
\par\vspace{3pt}
{\fontsize{7}{8}\selectfont\raggedright The \bestresult{} and \secondresult{} results are highlighted. All ALFWorld results in this table are on seen games (unseen: Table~\ref{tab:evoke-unseen}). ``-'' denotes unreported results. * denotes evaluation with skills.\par}
\end{table}

%% file: tables/results_alfworld_webshop.tex
\begingroup
\centering
\fontsize{6.2}{7.0}\selectfont
\setlength{\tabcolsep}{0.9pt}
\renewcommand{\arraystretch}{0.97}
\begin{tabular*}{\linewidth}{@{\extracolsep{\fill}}lrrrrrrr@{\hspace{3pt}}rr@{}}
\toprule
& \multicolumn{7}{c}{\textbf{ALFWorld}} & \multicolumn{2}{c}{\textbf{WebShop}} \\
\cmidrule(lr){2-8}\cmidrule(l){9-10}
\textbf{Method} & \textbf{Pick} & \textbf{Look} & \textbf{Clean} & \textbf{Heat} & \textbf{Cool} & \textbf{Pick2} & \textbf{Avg} & \textbf{Score} & \textbf{Succ.} \\
\midrule
\rowcolor{mainbackbone}
\multicolumn{10}{c}{\textit{Qwen2.5-3B-Instruct}} \\
\multicolumn{10}{@{}l}{\textit{Prompting}} \\
Vanilla & 44.4 & 11.1 & 6.2 & 15.4 & 28.6 & 12.5 & 19.7 & 6.7 & 0.8 \\[\improvedAWrowextra]
Skill-Prompt* & 51.7 & 66.7 & 48.4 & 0.0 & 4.3 & 10.0 & 30.2 & 0.2 & 0.8 \\[\improvedAWrowextra]
\multicolumn{10}{@{}l}{\textit{Policy optimization and distillation}} \\
OPSD & 48.8 & 41.7 & 16.7 & 0.0 & 15.8 & 16.7 & 23.3 & 11.3 & 3.1 \\[\improvedAWrowextra]
GRPO & 91.2 & 62.5 & 96.2 & 61.9 & 65.0 & 47.4 & 70.7 & 79.8 & 63.3 \\[\improvedAWrowextra]
Skill-GRPO & 88.9 & 71.4 & 58.8 & 70.6 & 40.7 & 29.2 & 59.9 & 77.3 & 60.9 \\[\improvedAWrowextra]
Skill-GRPO* & 94.3 & 57.1 & \improvedbest{100.0} & 66.7 & 73.1 & 57.1 & 74.7 & 76.3 & 66.4 \\[\improvedAWrowextra]
GRPO+OPSD & \improvedbest{100.0} & \improvedbest{82.4} & 85.7 & \improvedsecondblue{75.0} & 70.0 & 60.0 & 78.9 & 77.8 & 66.4 \\[\improvedAWrowextra]
Skill-SD & 88.2 & 50.0 & 96.2 & 52.4 & 65.0 & 57.9 & 68.3 & 75.9 & 64.0 \\[\improvedAWrowextra]
RLSD & 87.9 & 75.0 & 90.9 & \improvedsecondblue{75.0} & 73.1 & 68.4 & 78.4 & 84.4 & 66.4 \\[\improvedAWrowextra]
SDAR & \improvedsecondblue{97.1} & 62.5 & \improvedbest{100.0} & 61.9 & \improvedsecondblue{75.0} & \improvedsecondblue{84.2} & \improvedsecondblue{80.1} & \improvedsecondyellow{85.0} & \improvedsecondyellow{68.0} \\[\improvedAWrowextra]
\cmidrule(lr){1-10}
\textbf{\evoke{} (Ours)} & 91.4 & \improvedsecondblue{76.9} & \improvedsecondblue{96.3} & \improvedbest{100.0} & \improvedbest{92.0} & \improvedbest{91.7} & \improvedbest{91.4} & \improvedbest{87.4} & \improvedbest{82.8} \\[0pt]
\midrule
\rowcolor{mainbackbone}
\multicolumn{10}{c}{\textit{Qwen2.5-7B-Instruct}} \\
\multicolumn{10}{@{}l}{\textit{Prompting}} \\
Vanilla & 36.1 & 22.2 & 3.1 & 0.0 & 0.0 & 0.0 & 10.2 & 5.9 & 1.6 \\[\improvedAWrowextra]
Skill-Prompt* & 51.7 & 50.0 & 32.3 & 5.3 & 4.3 & 0.0 & 23.9 & 1.7 & 0.8 \\[\improvedAWrowextra]
\multicolumn{10}{@{}l}{\textit{Policy optimization and distillation}} \\
OPSD & 50.0 & 60.0 & 22.7 & 21.4 & 17.6 & 9.5 & 30.2 & 4.5 & 2.3 \\[\improvedAWrowextra]
GRPO & 91.2 & 87.5 & 96.2 & 81.0 & 65.0 & 57.9 & 79.8 & 80.9 & 72.6 \\[\improvedAWrowextra]
Skill-GRPO & 88.5 & 66.7 & 65.2 & 61.1 & 57.7 & 73.1 & 68.7 & 80.4 & 71.9 \\[\improvedAWrowextra]
Skill-GRPO* & \improvedbest{100.0} & 83.3 & 96.4 & 83.3 & 75.0 & 78.9 & 86.2 & 87.0 & 81.2 \\[\improvedAWrowextra]
GRPO+OPSD & 91.4 & 61.5 & \improvedbest{100.0} & 87.5 & 76.5 & 52.2 & 78.2 & 86.8 & 76.5 \\[\improvedAWrowextra]
Skill-SD & 93.9 & \improvedsecondblue{93.8} & 90.9 & \improvedbest{100.0} & 69.2 & 68.4 & 86.0 & 86.1 & 76.5 \\[\improvedAWrowextra]
RLSD & \improvedbest{100.0} & 87.5 & 92.3 & 58.8 & 80.0 & 65.2 & 80.6 & 87.4 & 77.3 \\[\improvedAWrowextra]
SDAR & 94.7 & 75.0 & \improvedbest{100.0} & 86.7 & 68.2 & 78.9 & 83.9 & \improvedsecondyellow{89.4} & \improvedsecondyellow{82.8} \\[\improvedAWrowextra]
\multicolumn{10}{@{}l}{\textit{World-model and dynamics methods}} \\
IWM & 90.6 & 42.9 & 85.2 & 88.2 & 84.2 & 76.9 & 78.0 & 69.5 & 56.2 \\[\improvedAWrowextra]
ITP$_{\mathrm R}$ & 94.3 & 76.0 & 88.9 & 87.5 & 53.8 & 91.7 & 82.0 & - & 60.2 \\[\improvedAWrowextra]
MemWM & 70.8 & 88.9 & 45.2 & 47.8 & 71.4 & 17.7 & 57.0 & 46.9 & - \\[\improvedAWrowextra]
GRPO+PaW & 90.4 & 80.7 & 86.8 & 82.9 & 76.5 & 67.3 & 80.8 & 84.5 & 70.5 \\[\improvedAWrowextra]
GIGPO+PaW & 98.2 & 85.6 & 98.6 & 84.5 & 91.5 & 84.3 & 90.5 & 87.6 & 76.7 \\[\improvedAWrowextra]
EnvRL-GRPO & 92.6 & 77.5 & 92.9 & 79.7 & 77.4 & 71.1 & 81.9 & 81.8 & 68.6 \\[\improvedAWrowextra]
EnvRL-GiGPO & \improvedsecondblue{98.3} & 92.2 & \improvedsecondblue{98.9} & \improvedsecondblue{93.8} & \improvedsecondblue{93.3} & \improvedsecondblue{94.6} & \improvedsecondblue{95.2} & 88.4 & 76.3 \\[\improvedAWrowextra]
\cmidrule(lr){1-10}
\textbf{\evoke{} (Ours)} & 94.3 & \improvedbest{100.0} & 96.3 & \improvedsecondblue{93.8} & \improvedbest{96.0} & \improvedbest{95.8} & \improvedbest{96.0} & \improvedbest{90.6} & \improvedbest{85.9} \\[0pt]
\ifdefined\improvedMeasurePrefix\else
\midrule
\rowcolor{mainbackbone}
\multicolumn{10}{c}{\textit{Qwen3-1.7B}} \\
\multicolumn{10}{@{}l}{\textit{Prompting}} \\
Vanilla & 25.0 & 22.2 & 3.1 & 0.0 & 21.4 & 4.2 & 12.7 & 46.5 & 4.7 \\[0pt]
Skill-Prompt* & 10.3 & 50.0 & 16.1 & 0.0 & 0.0 & 5.0 & 13.6 & 23.0 & 2.3 \\[0pt]
\multicolumn{10}{@{}l}{\textit{Policy optimization and distillation}} \\
OPSD & 26.3 & 33.3 & 9.1 & 0.0 & 4.5 & 5.3 & 13.1 & 47.4 & 9.3 \\[0pt]
GRPO & 71.1 & 41.7 & 36.4 & 40.0 & 31.8 & 31.6 & 42.1 & 67.3 & 38.3 \\[0pt]
Skill-GRPO & 27.6 & \improvedsecondblue{54.5} & 22.7 & 27.3 & 0.0 & 19.2 & 25.2 & 73.4 & 46.1 \\[0pt]
Skill-GRPO* & 31.4 & 42.9 & 51.9 & 8.3 & 11.5 & 7.1 & 25.5 & 80.4 & 50.0 \\[0pt]
GRPO+OPSD & 38.2 & 50.0 & 30.8 & 28.6 & 30.0 & 21.1 & 33.1 & 70.7 & 38.3 \\[0pt]
Skill-SD & 52.9 & 37.5 & 69.2 & \improvedsecondblue{42.9} & \improvedsecondblue{60.0} & \improvedsecondblue{36.8} & \improvedsecondblue{49.9} & \improvedsecondyellow{81.8} & 53.9 \\[0pt]
RLSD & 50.0 & 37.5 & 61.5 & 19.0 & 50.0 & 21.1 & 39.9 & 74.0 & 50.8 \\[0pt]
SDAR & \improvedsecondblue{73.5} & 25.0 & \improvedsecondblue{76.9} & 33.3 & 40.0 & \improvedsecondblue{36.8} & 47.6 & 76.8 & \improvedsecondyellow{58.6} \\[0pt]
\cmidrule(lr){1-10}
\textbf{\evoke{} (Ours)} & \improvedbest{94.3} & \improvedbest{84.6} & \improvedbest{96.3} & \improvedbest{87.5} & \improvedbest{92.0} & \improvedbest{91.7} & \improvedbest{91.1} & \improvedbest{88.4} & \improvedbest{80.5} \\[0pt]
\fi
\bottomrule
\end{tabular*}
\par\endgroup

%% file: tables/results_search_qa.tex
\begingroup
\centering
\fontsize{6.2}{7.0}\selectfont
\setlength{\tabcolsep}{0.9pt}
\renewcommand{\arraystretch}{0.97}
\begin{tabular*}{\linewidth}{@{\extracolsep{\fill}}lrrrrrrrr@{}}
\toprule
& \multicolumn{8}{c}{\textbf{Search-based QA}} \\
\cmidrule(l){2-9}
\textbf{Method} & \textbf{NQ} & \textbf{Triv} & \textbf{Pop} & \textbf{Hotp} & \textbf{2Wk} & \textbf{MuS} & \textbf{Bam} & \textbf{Avg} \\
\midrule
\rowcolor{mainbackbone}
\multicolumn{9}{c}{\textit{Qwen2.5-3B-Instruct}} \\
\multicolumn{9}{@{}l}{\textit{Prompting}} \\
Vanilla & 24.6 & 48.1 & 31.0 & 26.3 & 25.3 & 7.2 & 59.7 & 31.7 \\[\improvedQArowextra]
Skill-Prompt* & 23.7 & 46.2 & 30.6 & 24.4 & 22.1 & 7.5 & 12.5 & 23.9 \\[\improvedQArowextra]
\multicolumn{9}{@{}l}{\textit{Policy optimization and distillation}} \\
OPSD & 0.1 & 0.1 & 0.1 & 0.0 & 0.0 & 0.0 & 0.0 & 0.0 \\[\improvedQArowextra]
GRPO & 39.3 & 60.6 & 41.1 & 37.4 & 34.6 & 15.4 & 26.4 & 36.4 \\[\improvedQArowextra]
Skill-GRPO & 43.5 & 58.8 & 43.0 & 36.8 & 32.2 & 11.7 & 12.5 & 34.1 \\[\improvedQArowextra]
Skill-GRPO* & 44.3 & 59.6 & 44.3 & 39.0 & 36.1 & 14.5 & 14.9 & 36.1 \\[\improvedQArowextra]
GRPO+OPSD & 44.9 & \improvedsecondgreen{61.2} & 45.2 & \improvedsecondgreen{40.4} & 38.5 & \improvedsecondgreen{16.0} & \improvedsecondgreen{66.1} & 44.6 \\[\improvedQArowextra]
Skill-SD & 44.4 & 60.4 & 44.0 & 39.5 & \improvedsecondgreen{40.4} & 15.4 & 64.9 & 44.1 \\[\improvedQArowextra]
RLSD & 41.5 & 58.6 & 42.3 & \improvedsecondgreen{40.4} & 40.2 & \improvedbest{16.8} & \improvedbest{66.9} & 43.8 \\[\improvedQArowextra]
SDAR & 44.8 & 58.1 & 44.3 & 38.6 & 36.2 & 15.7 & \improvedsecondgreen{66.1} & 43.4 \\[\improvedQArowextra]
\multicolumn{9}{@{}l}{\textit{World-model and dynamics methods}} \\
GRPO+PaW & 45.8 & \improvedsecondgreen{61.2} & \improvedbest{47.5} & 39.4 & 40.1 & 14.4 & 65.2 & \improvedsecondgreen{44.8} \\[\improvedQArowextra]
GIGPO+PaW & \improvedbest{46.2} & \improvedbest{61.8} & \improvedsecondgreen{46.7} & 37.6 & 38.0 & 13.9 & 64.9 & 44.2 \\[\improvedQArowextra]
\cmidrule(lr){1-9}
\textbf{\evoke{} (Ours)} & \improvedsecondgreen{46.1} & \improvedbest{61.8} & 45.2 & \improvedbest{41.5} & \improvedbest{41.0} & 15.3 & \improvedsecondgreen{66.1} & \improvedbest{45.3} \\[0pt]
\midrule
\rowcolor{mainbackbone}
\multicolumn{9}{c}{\textit{Qwen2.5-7B-Instruct}} \\
\multicolumn{9}{@{}l}{\textit{Prompting}} \\
Vanilla & 25.2 & 50.8 & 29.5 & 29.0 & 29.0 & 10.4 & 63.7 & 33.9 \\[\improvedQArowextra]
Skill-Prompt* & 30.9 & 52.1 & 32.7 & 32.7 & 27.9 & 12.7 & 66.1 & 36.4 \\[\improvedQArowextra]
\multicolumn{9}{@{}l}{\textit{Policy optimization and distillation}} \\
OPSD & 8.8 & 8.6 & 17.5 & 2.5 & 4.2 & 0.5 & 1.2 & 6.2 \\[\improvedQArowextra]
GRPO & 45.1 & 63.7 & 44.0 & 43.6 & 43.2 & 16.8 & 37.6 & 42.0 \\[\improvedQArowextra]
Skill-GRPO & 45.2 & 63.7 & 45.7 & 43.1 & 43.3 & 19.6 & 21.4 & 40.3 \\[\improvedQArowextra]
Skill-GRPO* & 44.8 & 63.0 & 45.1 & 43.7 & 43.7 & \improvedsecondgreen{20.5} & \improvedsecondgreen{71.4} & 47.5 \\[\improvedQArowextra]
GRPO+OPSD & \improvedsecondgreen{47.3} & 64.5 & 46.9 & 43.8 & 39.3 & 18.0 & 69.4 & 47.0 \\[\improvedQArowextra]
Skill-SD & 47.1 & 64.5 & 47.8 & 44.2 & 42.1 & 20.2 & 69.0 & 47.8 \\[\improvedQArowextra]
RLSD & 46.8 & 63.0 & 44.4 & \improvedbest{45.5} & \improvedsecondgreen{48.9} & \improvedbest{21.5} & \improvedbest{73.0} & \improvedsecondgreen{49.0} \\[\improvedQArowextra]
SDAR & 46.3 & 63.5 & \improvedsecondgreen{48.2} & 43.8 & 48.4 & 19.6 & \improvedbest{73.0} & \improvedsecondgreen{49.0} \\[\improvedQArowextra]
\multicolumn{9}{@{}l}{\textit{World-model and dynamics methods}} \\
GRPO+PaW & \improvedbest{48.9} & 64.9 & \improvedbest{48.5} & \improvedsecondgreen{44.9} & 45.1 & 18.9 & 70.1 & 48.8 \\[\improvedQArowextra]
GIGPO+PaW & 46.5 & \improvedbest{66.0} & 47.2 & 42.2 & 42.8 & 18.6 & 69.5 & 47.5 \\[\improvedQArowextra]
\cmidrule(lr){1-9}
\textbf{\evoke{} (Ours)} & 46.8 & \improvedsecondgreen{65.5} & 47.8 & 44.5 & \improvedbest{49.4} & 20.0 & 70.5 & \improvedbest{49.2} \\[0pt]
\ifdefined\improvedMeasurePrefix\else
\midrule
\rowcolor{mainbackbone}
\multicolumn{9}{c}{\textit{Qwen3-1.7B}} \\
\multicolumn{9}{@{}l}{\textit{Prompting}} \\
Vanilla & 29.4 & 46.9 & 37.0 & 23.5 & 19.6 & 6.4 & 10.5 & 24.8 \\[0pt]
Skill-Prompt* & 29.4 & 46.5 & 36.2 & 22.9 & 20.8 & 4.3 & 10.1 & 24.3 \\[0pt]
\multicolumn{9}{@{}l}{\textit{Policy optimization and distillation}} \\
OPSD & 4.2 & 8.3 & 4.6 & 6.6 & 15.3 & 0.7 & 1.2 & 5.8 \\[0pt]
GRPO & 40.0 & \improvedsecondgreen{58.9} & 43.5 & 35.4 & 30.3 & 12.0 & 65.7 & 40.8 \\[0pt]
Skill-GRPO & 39.2 & 58.6 & 43.9 & 35.2 & 28.2 & 11.5 & \improvedsecondgreen{66.1} & 40.4 \\[0pt]
Skill-GRPO* & 38.0 & 58.4 & 43.9 & 36.3 & 29.0 & 12.5 & \improvedbest{66.9} & 40.7 \\[0pt]
GRPO+OPSD & \improvedsecondgreen{40.7} & \improvedsecondgreen{58.9} & 45.0 & \improvedsecondgreen{37.0} & 34.6 & \improvedsecondgreen{13.3} & 65.7 & \improvedsecondgreen{42.2} \\[0pt]
Skill-SD & 39.1 & 57.5 & \improvedbest{45.4} & 34.8 & 34.1 & 10.7 & 64.1 & 40.8 \\[0pt]
RLSD & 38.6 & 57.3 & 43.0 & 34.5 & 34.1 & 11.5 & 65.3 & 40.6 \\[0pt]
SDAR & 39.7 & \improvedsecondgreen{58.9} & \improvedsecondgreen{45.3} & 35.9 & \improvedsecondgreen{35.5} & 12.6 & 65.3 & 41.9 \\[0pt]
\cmidrule(lr){1-9}
\textbf{\evoke{} (Ours)} & \improvedbest{44.5} & \improvedbest{60.4} & 43.3 & \improvedbest{41.1} & \improvedbest{39.6} & \improvedbest{14.2} & 64.1 & \improvedbest{43.9} \\[0pt]
\fi
\bottomrule
\end{tabular*}
\par\endgroup

%% file: tables/evoke_alfworld_unseen.tex
\begin{table}[!t]
\centering\fontsize{8}{8.4}\selectfont
\setlength{\aboverulesep}{1pt}\setlength{\belowrulesep}{1pt}
\setlength{\tabcolsep}{3pt}
\setlength{\abovecaptionskip}{0pt}
\caption{ALFWorld Unseen results (\%). The \bestresult{} and \secondresult{} results are highlighted within each backbone.}
\label{tab:evoke-unseen}
\begin{tabular*}{\linewidth}{@{\extracolsep{\fill}}lrrrrrrr@{}}
\toprule
\textbf{Method} & \textbf{Pick} & \textbf{Look} & \textbf{Clean} & \textbf{Heat} & \textbf{Cool} & \textbf{Pick2} & \textbf{Avg} \\
\midrule
\rowcolor{mainbackbone}\multicolumn{8}{c}{\textit{Qwen2.5-3B-Instruct}} \\
\multicolumn{8}{@{}l}{\textit{Prompting}} \\
ReAct & 17.4 & 6.7 & 8.8 & 7.4 & 9.1 & 0.0 & 8.2 \\
\multicolumn{8}{@{}l}{\textit{Policy optimization and distillation}} \\
GRPO & 73.9 & 60.0 & 82.4 & 59.3 & 72.7 & \improvedsecondblue{76.9} & 70.9 \\
SDAR & 71.0 & 77.5 & 71.8 & 73.9 & 73.2 & 57.6 & 70.8 \\
OPID & 78.3 & \improvedbest{86.7} & 82.4 & 77.8 & 77.3 & 69.2 & 78.6 \\
PCSD & \improvedsecondblue{95.1} & \improvedsecondblue{80.9} & \improvedsecondblue{89.6} & 81.5 & \improvedsecondblue{92.4} & 75.9 & \improvedsecondblue{85.9} \\
AHEAD & 83.3 & 77.8 & 77.4 & \improvedsecondblue{87.0} & 76.2 & \improvedbest{94.1} & 82.6 \\
\cmidrule(lr){1-8}
\textbf{\evoke{} (Ours)} & \improvedbest{95.8} & 72.2 & \improvedbest{93.6} & \improvedbest{95.7} & \improvedbest{95.2} & \improvedbest{94.1} & \improvedbest{91.1} \\
\midrule
\rowcolor{mainbackbone}\multicolumn{8}{c}{\textit{Qwen2.5-7B-Instruct}} \\
\multicolumn{8}{@{}l}{\textit{Policy optimization and distillation}} \\
GRPO & 62.5 & 38.9 & \improvedsecondblue{83.9} & 65.2 & \improvedsecondblue{85.7} & \improvedsecondblue{52.9} & 64.9 \\
AHEAD & \improvedsecondblue{79.2} & \improvedbest{83.3} & \improvedsecondblue{83.9} & \improvedsecondblue{82.6} & \improvedsecondblue{85.7} & \improvedbest{88.2} & \improvedsecondblue{83.8} \\
\multicolumn{8}{@{}l}{\textit{World-model and dynamics methods}} \\
IWM & - & - & - & - & - & - & 70.3 \\
EnvRL-GRPO & - & - & - & - & - & - & 70.2 \\
EnvRL-GiGPO & - & - & - & - & - & - & 81.8 \\
\cmidrule(lr){1-8}
\textbf{\evoke{} (Ours)} & \improvedbest{95.8} & \improvedsecondblue{77.8} & \improvedbest{96.8} & \improvedbest{100.0} & \improvedbest{100.0} & \improvedbest{88.2} & \improvedbest{93.1} \\
\midrule
\rowcolor{mainbackbone}\multicolumn{8}{c}{\textit{Qwen3-1.7B}} \\
\multicolumn{8}{@{}l}{\textit{Policy optimization and distillation}} \\
GRPO & 79.2 & 66.7 & \improvedsecondblue{77.4} & 87.0 & 71.4 & 64.7 & 74.4 \\
SDAR & 75.0 & \improvedsecondblue{72.2} & \improvedsecondblue{77.4} & 87.0 & 76.2 & 76.5 & 77.4 \\
GRSD & \improvedsecondblue{83.3} & \improvedbest{77.9} & \improvedsecondblue{77.4} & \improvedsecondblue{91.3} & \improvedbest{85.7} & \improvedsecondblue{88.2} & \improvedsecondblue{84.0} \\
AHEAD & 66.7 & 44.4 & 51.6 & 73.9 & 57.1 & 58.8 & 58.8 \\
\cmidrule(lr){1-8}
\textbf{\evoke{} (Ours)} & \improvedbest{87.5} & 55.6 & \improvedbest{93.6} & \improvedbest{95.7} & \improvedsecondblue{81.0} & \improvedbest{94.1} & \improvedbest{84.6} \\
\bottomrule
\end{tabular*}
\end{table}

%% file: sections/5_analysis.tex
\section{Ablations and Analysis}
\label{sec:analysis}
We examine where the gains of \evoke{} come from and how they arise. All analyses use Qwen2.5-3B on ALFWorld. Every variant starts from the same walkthrough-initialized policy $\pi_{\mathrm{boot}}$ and, unless noted, uses the same preference data and training schedule as \evoke{}. Each checkpoint is evaluated on the 140 seen and 134 unseen games with greedy, free-form action generation, and we report micro success averaged over training seeds (Appendix~\ref{app:statistics}).

\subsection{Do the gains come from goal interventions?}
\label{sec:supervision-analysis}
\emph{Source only} in Table~\ref{tab:controlled} trains with the same ranking objective as \evoke{} but only on the original goals, removing every alternative-goal context. Unseen success drops from 91.8\% to 82.8\%, and the policy needs 40.7\% more actions per game. This is not a matter of training longer: \emph{Source replay} repeats the original-goal data until it matches the updates of \evoke{}, and still reaches only 85.9\% while taking 34.4\% more actions. The gain persists under a fixed supervision budget and across paired training seeds (Appendix~\ref{app:extra-ablations}).\looseness=-1

\begin{table}[!htbp]
\centering\small\setlength{\tabcolsep}{5pt}
\caption{\textbf{Decision-supervision ablations.} Micro success (\%) and average unseen steps, averaged over training seeds. All variants start from $\pi_{\mathrm{boot}}$. The \bestresult{} and \secondresult{} results are highlighted.}
\label{tab:controlled}
\begin{tabular}{lccrrr}
\toprule
\textbf{Method} & \textbf{Alt. goals} & \textbf{Objective} & \textbf{\boldmath Seen $\uparrow$} & \textbf{\boldmath Unseen $\uparrow$} & \textbf{\boldmath Avg Steps $\downarrow$} \\
\midrule
$\pi_{\mathrm{boot}}$ & --- & Walkthrough SFT & 69.3 & 60.4 & 25.9 \\
Source only & No & Ranking & 82.9 & 82.8 & 16.0 \\
Source replay & No & Ranking & 90.4 & \cellcolor{evokesecond}\underline{85.9} & 15.3 \\
Positive SFT & Yes & SFT & \cellcolor{evokesecond}\underline{91.4} & 84.3 & \cellcolor{evokesecond}\underline{15.0} \\
\evoke{} & Yes & Ranking & \cellcolor{evokebest}\textbf{92.1} & \cellcolor{evokebest}\textbf{91.8} & \cellcolor{evokebest}\textbf{11.4} \\
\bottomrule
\end{tabular}
\end{table}

\subsection{Is more goal data enough?}
\label{sec:ranking-analysis}
Goal interventions also multiply the number of goals the policy is trained on, so the gain could simply reflect more and more diverse training tasks. \emph{Positive SFT} receives exactly the same goal contexts, positive actions, and updates as \evoke{}, but imitates the positives instead of ranking them. It nearly matches \evoke{} on seen games but falls 7.5 points behind on unseen games. More goal data with imitation fits the training environments but transfers much less, which is the surface-level behavior matching discussed in Section~\ref{sec:intro}. Imitation raises the positives without directly pushing down the policy's habitual choices; ranking places these choices below the action each goal calls for, and transfers best when the competitors are the actions the policy itself favors (Appendix~\ref{app:extra-ablations}).

\begin{figure}[!htbp]
\setlength{\abovecaptionskip}{6pt}
\begin{minipage}[t]{.49\linewidth}
\centering
\includegraphics[width=\linewidth]{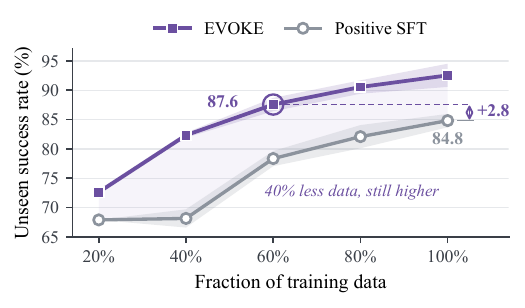}
\caption{\textbf{Data efficiency.} \evoke{} outperforms SFT at every budget. Unseen success vs.\ fraction of labeled states; mean $\pm$ s.d.\ over three training orders.}
\label{fig:state-budget}
\end{minipage}\hfill
\begin{minipage}[t]{.49\linewidth}
\centering
\includegraphics[width=\linewidth]{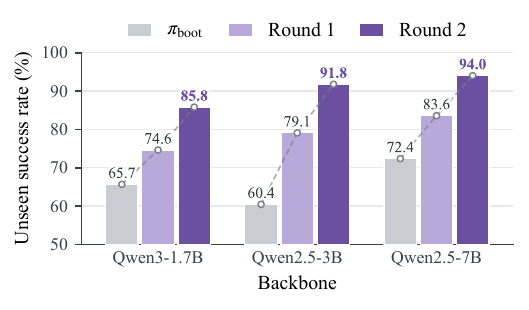}
\caption{\textbf{Iterative improvement.} Every round improves every backbone. Unseen success across training rounds for three backbones.}
\label{fig:evolution}
\end{minipage}
\end{figure}

The advantage holds at every amount of training data (Figure~\ref{fig:state-budget}; nested subsets of 20--100\% of the labeled states, three training orders each). \evoke{} outperforms SFT in all 15 paired runs; with 60\% of the states, it already exceeds SFT trained on all of them. Ranking thus extracts more from each labeled state.

\subsection{Does iteration keep improving the policy?}
\label{sec:learning-analysis}
Each round lets the updated policy reach new states and expose new mistakes, which become new hard negatives. Figure~\ref{fig:evolution} follows three backbones through two rounds of aggregation. Unseen success rises by 20.1--31.3 points on every backbone, with a clear gain in every round (Appendix Table~\ref{tab:lineage}). On Qwen2.5-3B, the two rounds together supervise about 9.1\% as many decisions as the demonstrations behind $\pi_{\mathrm{boot}}$ (Appendix Table~\ref{tab:supervision}).

\begin{figure}[!b]
\setlength{\abovecaptionskip}{6pt}
\centering
\includegraphics[width=\linewidth]{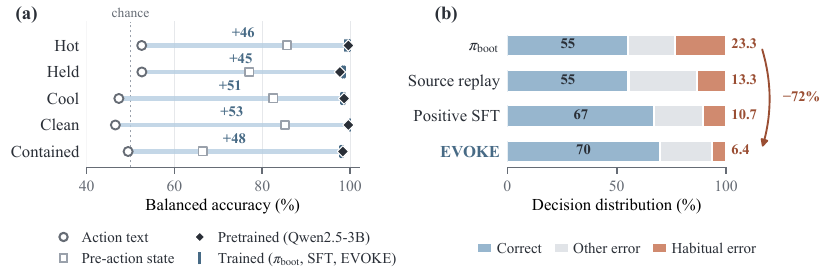}
\caption{\textbf{From knowledge to goal-directed decisions on unseen games.} Consequences are decodable before any ALFWorld training, and \evoke{} makes the fewest habitual errors. (a) Balanced accuracy of linear probes that predict an action's consequence from hidden states before the action is executed, on actions whose text leads to different outcomes; Qwen2.5-3B is the backbone before any ALFWorld training. (b) Outcome of each decision on 812 goal pairs that require different actions. A \emph{habitual error} chooses the action that is correct for the other goal of the pair.}
\label{fig:knowledge-use}
\end{figure}

\subsection{Is the knowledge already there?}
\label{sec:knowledge-analysis}
If \evoke{} elicits rather than adds knowledge, the pretrained model should already encode what its actions do. We feed each model a goal, a history, and a candidate action, extract its hidden state before the action is executed, and train a linear probe to predict five consequences observed after execution: whether the target object is held, lies in the receptacle named by the action, or is hot, clean, or cool. Probes are trained on training games, with layer and regularization selected on held-out training games, and tested on unseen games (Appendix~\ref{app:probe}).

Figure~\ref{fig:knowledge-use}a shows that consequences are almost perfectly decodable in every model, including the original Qwen2.5-3B backbone before any ALFWorld training. This is not a surface cue: on transitions where the same action text leads to different outcomes, a probe on the action text is at chance, while the model representations remain above 97.0\%, against 65.0--68.0\% for the same architecture with random weights (Appendix Table~\ref{tab:probe-controls}). Yet $\pi_{\mathrm{boot}}$, whose representations are equally decodable, succeeds on only 60.4\% of unseen games, and \evoke{} raises this to 91.8\% (Table~\ref{tab:controlled}). The backbone already represents what its actions do; what remains is whether the policy uses this to decide.\looseness=-1

\subsection{Does \evoke{} turn this knowledge into goal-directed decisions?}
\label{sec:goal-analysis}
A policy relying on contextual habits cannot choose correctly when the same context calls for different actions under different goals (Section~\ref{sec:goal-intervention}). We test this directly: from the 134 unseen games, we construct 812 goal pairs in which two goals share the same state, history, and available actions but have disjoint sets of progress-making actions. Each model scores all available actions under both goals, and a pair counts as correct only if the top-ranked action is a positive under each goal; random choice succeeds on 0.3\% of pairs. We also count \emph{habitual errors}: decisions that choose the action correct for the other goal of the pair (Appendix~\ref{app:goal-swap}).

Figure~\ref{fig:knowledge-use}b makes the habit visible. $\pi_{\mathrm{boot}}$ solves only 19.5\% of pairs, and more than half of its errors are habitual: it follows the context rather than the goal. Training on the intervened data more than doubles joint accuracy, and ranking suppresses the habit further. \evoke{} makes the fewest habitual errors (6.4\% of decisions), 4.3 points fewer than Positive SFT (95\% CI [3.0, 5.6]). Ranking alone already helps, since Source replay reduces habitual errors even without alternative goals, and combined with goal interventions it reduces them the most. \evoke{} thus decides by the goal rather than by habit.\looseness=-1

\begingroup
\setlength{\intextsep}{12pt plus 2pt minus 2pt}
\begin{figure}[!h]
\setlength{\abovecaptionskip}{6pt}
\centering
\includegraphics[width=\linewidth]{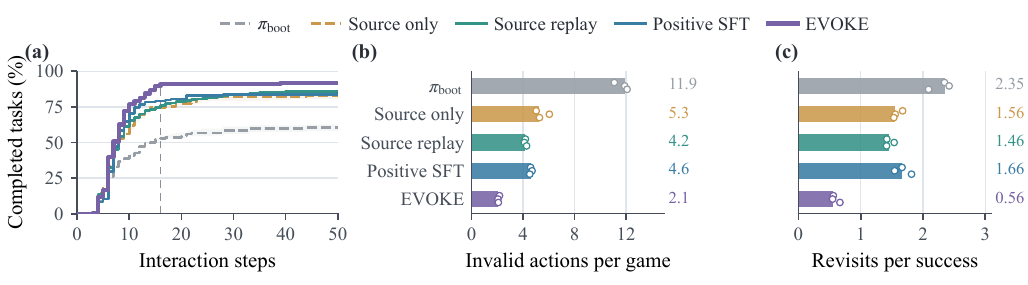}
\caption{\textbf{Execution on unseen games.} \evoke{} succeeds earlier and wastes fewer actions. (a) Cumulative success over interaction steps. (b) Invalid actions per game. (c) Revisits to an already visited location per successful game.}
\label{fig:execution}
\end{figure}
\endgroup

\subsection{Better decisions or more trial and error?}
\label{sec:execution-analysis}
With free-form actions and a 50-step budget, higher success could come from persistence rather than better decisions. Figure~\ref{fig:execution} separates the two. \evoke{} solves 91.0\% of unseen games within 20 actions, versus at most 80.6\% for the other trained variants, and needs 24.4--28.9\% fewer actions per game (Table~\ref{tab:controlled}). The speed-up is not merely a consequence of failing less: on the 70 games that all five variants solve, \evoke{} still takes the fewest actions. It also makes less than half as many invalid actions as Positive SFT and about a third as many revisits as the other trained variants. \evoke{} thus reaches the goal along more direct paths instead of trying more (examples in Appendix Table~\ref{tab:qualitative-decisions}).

\FloatBarrier

%% file: sections/6_conclusion.tex
\section{Conclusion}
\label{sec:conclusion}
\vspace{1mm}
We presented \evoke{}, which elicits the world knowledge of pretrained LLM agents for transferable decisions by ranking actions under different goals at fixed states, against the policy's own tempting mistakes. Across household tasks, web navigation, and search-based QA with three backbones, \evoke{} improves task performance and generalization to unseen environments. Our analyses trace the gains to goal interventions and ranking, and indicate that the pretrained backbone already represents what its actions do, and that \evoke{} makes the policy decide by the goal rather than by habit. For LLM agents, eliciting world knowledge through decision supervision offers a practical alternative to acquiring it through prediction.\looseness=-1

%% file: statements.tex
\subsection*{AI Use Statement}
Large language models (LLMs) were used to aid in the writing and polishing of the manuscript. Within the method, an LLM assists in data construction: it proposes alternative goals and assesses executed candidate actions to build preference labels (Section~\ref{sec:construction} and Appendix~\ref{app:labeling}). The annotator is not used at deployment.
All research ideas, methods, and conclusions were developed by the authors. The authors bear full responsibility for the content of the manuscript, including any text generated or polished by an LLM, and have ensured that it complies with ethical guidelines and does not involve plagiarism or scientific misconduct.

\subsection*{Reproducibility Statement}
We have made every effort to ensure that our results are reproducible. Section~\ref{sec:method} and Appendix~\ref{app:labeling} describe preference construction, the training objective, model architecture, and hyperparameters; Appendix~\ref{app:statistics} describes the evaluation protocol, statistical aggregation, and data-efficiency subsets; Appendix~\ref{app:probe} and Appendix~\ref{app:goal-swap} detail the diagnostic analyses. All benchmarks used in our experiments are publicly available. We will release the source code, trained checkpoints, and experiment scripts to support independent verification and extension of this work.

%% file: appendix.tex
\makeatletter
\newenvironment{appendixtable}{%
  \par\addvspace{10pt}\noindent
  \begin{minipage}{\linewidth}%
  \def\@captype{table}%
  \setlength{\abovecaptionskip}{0pt}%
  \setlength{\belowcaptionskip}{5pt}%
}{%
  \end{minipage}\par\addvspace{10pt}%
}
\makeatother
\raggedbottom

\section{Implementation Details of \evoke{}}
\label{app:labeling}
This section follows the order of Section~\ref{sec:method}.

\paragraph{Policy input and demonstration initialization.}
The policy input consists of the goal, the visible action--observation history, and the list of available actions; the output is the next action as text. $\pi_{\mathrm{boot}}$ is obtained by supervised fine-tuning on 16,247 per-step examples from successful demonstrations in 2,700 training games, for one epoch with learning rate $2\!\times\!10^{-4}$, batch size 2, gradient accumulation 8, warmup ratio 0.03, and maximum length 1,280 tokens. LoRA adapters of rank 16 and scaling 16 are attached to the attention query, key, value, and output projections and to the feed-forward gate, up, and down projections.

\paragraph{Decision-state collection.}
The policy runs in training games, and each step records the goal and the visible history, so that the state can be revisited. States are sampled along the trajectories, covering both steps that make progress and steps that precede detours, repeated actions, or failure. The first round yields 344 states from 193 games; after aggregation, the dataset contains 767 states from 312 games (Table~\ref{tab:training}). The two rounds supervise 1,482 decisions, about 9.1\% of the 16,247 demonstration steps used for $\pi_{\mathrm{boot}}$, and raise unseen success from 60.4\% to 91.8\% (Table~\ref{tab:supervision}).

\paragraph{Goal interventions.}
Given the scene and the original goal, the LLM annotator (Qwen2.5-72B-Instruct) proposes alternative goals. Each proposal is instantiated in the environment together with its success condition. A goal is kept if it is compatible with the scene, not yet satisfied, and achievable from the current state. The original task text is removed from the visible history, so the policy sees only the goal of the current context, while the environment state, history, and available actions are shared by all goals of a state.

\paragraph{Candidate actions and annotation.}
For each state--goal context, candidates are drawn from $\mathcal A(s)$: the policy's highest-scoring actions, further available actions, and annotator suggestions that match an available action. Each candidate is executed once from the same state, and its resulting observation is recorded, followed by a short continuation when a single step does not reveal the effect of the action. The annotator receives the goal, the history before the action, and every candidate paired with its executed outcome, and classifies each candidate as advancing the goal, worse, or uncertain. Advancing actions form $P$, worse actions form the competitor pool, and uncertain actions are discarded. The annotator thus serves as a labeler of observed outcomes rather than a planner: each label rests on an executed transition and a goal whose success condition is defined by the environment.

\paragraph{Preference groups and hard negatives.}
Each group contains up to four positives and four competitors. Competitors are selected in order of priority: the policy's greedy action when it is not a positive, then non-positive actions with the highest mean log-probability under the policy, and then actions of other types for coverage. A context with more than four positives is split into several groups that share the same competitors, and each group is weighted by the inverse of the number of groups in its context. In the final dataset, every group has four competitors and on average 1.6 positives.

\paragraph{Optimization.}
Action scores average the token log-probabilities of the action completion, including its end-of-turn token. The ranking loss uses $\tau=1$ and $\lambda=0.25$. Each round starts a new AdamW optimizer with zero weight decay, continues the LoRA adapter from the previous round, and accumulates gradients over eight groups; the backbone stays frozen throughout. The final checkpoint of each round is used. Table~\ref{tab:training} lists the schedule. The positive-SFT variant in Section~\ref{sec:analysis} uses the same groups, weights, and schedule, and minimizes $\mathcal L_{\mathrm{SFT}}=-|P|^{-1}\sum_{p\in P}z_\theta(p\mid x_g)$.

\begin{appendixtable}
\centering\small
\caption{\textbf{Training schedule of the analyses in Section~\ref{sec:analysis}.} Round~2 trains on the aggregate of both rounds.}
\label{tab:training}
\begin{tabular}{lrrrrrrr}
\toprule
\textbf{Round} & \textbf{States} & \textbf{Contexts} & \textbf{Groups} & \textbf{Games} & \textbf{Epochs} & \textbf{Learning rate} & \textbf{Updates} \\
\midrule
1 & 344 & 687 & 706 & 193 & 2 & $10^{-5}$ & 178 \\
2 (aggregate) & 767 & 1,482 & 1,524 & 312 & 1 & $5\!\times\!10^{-6}$ & 191 \\
\bottomrule
\end{tabular}
\end{appendixtable}

\begin{appendixtable}
\centering\small
\caption{\textbf{Supervision and success.} $\pi_{\mathrm{boot}}$ imitates one action per demonstration step; \evoke{} ranks the candidate actions of each context. Micro success (\%).}
\label{tab:supervision}
\begin{tabular}{lrrrrr}
\toprule
& \textbf{Supervised decisions} & \textbf{\boldmath \% of $\pi_{\mathrm{boot}}$} & \textbf{Games} & \textbf{Seen} & \textbf{Unseen} \\
\midrule
$\pi_{\mathrm{boot}}$ & 16,247 & 100.0 & 2,700 & 69.3 & 60.4 \\
+ Round 1 & 687 & 4.2 & 193 & 88.6 & 79.1 \\
+ Round 2 (aggregate) & 1,482 & 9.1 & 312 & 92.1 & 91.8 \\
\bottomrule
\end{tabular}
\end{appendixtable}

\paragraph{Other environments.}
The same pipeline applies to WebShop and search-based QA; only the form of goals, states, and actions changes. In WebShop, for example, a goal can be a purchase instruction, a state the current page with its history, and a candidate action a search query or a click. In search-based QA, a goal can be a question, a state the retrieval history so far, and a candidate action a new query or an answer; an alternative goal is then another question the same retrieval history can lead to, such as one about an entity already retrieved. In both environments, the annotator labels the executed outcomes, i.e., the next page or the retrieved documents.

\section{Evaluation Protocol}
\label{app:statistics}
\paragraph{Evaluation.}
ALFWorld Seen and Unseen contain 140 and 134 games, with 35/13/27/16/25/24 and 24/18/31/23/21/17 games in Pick/Look/Clean/Heat/Cool/Pick2 order. At each step, the policy receives the goal, the full visible history, and the list of available actions, and generates the next action greedily as free text; no legal-action masking or output repair is applied, and the output is passed to the environment as is. An episode succeeds if the environment reports task completion within 50 actions.

\paragraph{Aggregation.}
Micro success is the fraction of successful games, and macro success is the unweighted mean over the six task types. Each checkpoint is evaluated with three seeds and summarized by the median, and all reported results are averaged over training seeds. The data-efficiency curves average three training orders per point. Confidence intervals resample games with replacement 10,000 times, pooling training seeds and keeping all observations of a game together.

\subsection{Data-Efficiency Protocol}
\label{app:data-budget}
Subsets contain 20, 40, 60, 80, or 100\% of the labeled states and are nested, so each smaller subset is contained in every larger one. States are stratified by collection round and task type, and every selected state keeps all of its goal contexts and preference groups. Both objectives use the same subsets, the same two-round schedule as the full model (Table~\ref{tab:training}), and the same three training orders, so the number of updates scales with the amount of data. Table~\ref{tab:data-budget} lists the subset sizes and Table~\ref{tab:budget-results} the results.

\begin{appendixtable}
\centering\small
\caption{\textbf{Data subsets.} States and goal contexts after aggregation.}
\label{tab:data-budget}
\begin{tabular}{lrrrrr}
\toprule
\textbf{Data} & \textbf{20\%} & \textbf{40\%} & \textbf{60\%} & \textbf{80\%} & \textbf{100\%} \\
\midrule
States & 154 & 307 & 460 & 613 & 767 \\
Contexts & 301 & 592 & 889 & 1,185 & 1,482 \\
\bottomrule
\end{tabular}
\end{appendixtable}

\begin{appendixtable}
\begin{minipage}[t]{\linewidth}
\centering\footnotesize\setlength{\tabcolsep}{3pt}
\caption{\textbf{Data efficiency.} Micro success (\%), mean $\pm$ s.d.\ over three training orders; all budgets, including 100\%, are retrained on nested subsets, independently of Table~\ref{tab:controlled}. $\Delta$: Unseen \evoke{} minus SFT.}
\label{tab:budget-results}
\begin{tabular}{lrrrrr}
\toprule
& \multicolumn{2}{c}{\textbf{Seen}} & \multicolumn{3}{c}{\textbf{Unseen}}\\
\cmidrule(lr){2-3}\cmidrule(lr){4-6}
\textbf{Data} & \textbf{\evoke{}} & \textbf{SFT} & \textbf{\evoke{}} & \textbf{SFT} & \textbf{\boldmath $\Delta$}\\
\midrule
20\% & 77.4$\pm$1.6 & 75.7$\pm$0.7 & 72.6$\pm$0.4 & 67.9$\pm$0.0 & +4.7\\
40\% & 82.4$\pm$1.5 & 78.6$\pm$0.0 & 82.3$\pm$0.4 & 68.2$\pm$1.6 & +14.2\\
60\% & 89.3$\pm$0.7 & 84.3$\pm$0.7 & 87.6$\pm$1.1 & 78.4$\pm$1.3 & +9.2\\
80\% & 92.1$\pm$1.9 & 85.7$\pm$0.0 & 90.5$\pm$1.1 & 82.1$\pm$2.0 & +8.5\\
100\% & 93.6$\pm$1.4 & 89.5$\pm$0.8 & 92.5$\pm$2.0 & 84.8$\pm$1.1 & +7.7\\
\bottomrule
\end{tabular}
\end{minipage}
\end{appendixtable}

\Needspace{100pt}
\section{Additional Results}
\label{app:ablation-details}

\begin{appendixtable}
\centering\small
\caption{\textbf{Iterative training.} Seen / Unseen micro success (\%) after each round.}
\label{tab:lineage}
\begin{tabular}{lrrr}
\toprule
\textbf{Backbone} & \textbf{\boldmath $\pi_{\mathrm{boot}}$} & \textbf{Round 1} & \textbf{Round 2}\\
\midrule
1.7B & 80.0 / 65.7 & 87.1 / 74.6 & 92.1 / 85.8 \\
3B & 69.3 / 60.4 & 88.6 / 79.1 & 92.1 / 91.8 \\
7B & 81.4 / 72.4 & 87.1 / 83.6 & 95.7 / 94.0 \\
\bottomrule\end{tabular}\end{appendixtable}

\input{tables/controlled_task_breakdown}

\Needspace{100pt}
\subsection{Additional Ablations}
\label{app:extra-ablations}
Both comparisons start from $\pi_{\mathrm{boot}}$ and follow the protocol of Section~\ref{sec:analysis}.

\paragraph{Alternative goals under a fixed budget.}
Two variants share the same 600 policy-visited states, 300 and 600 contexts in the two rounds, the ranking objective and competitor selection of \evoke{}, and 76 + 75 updates. They differ only in whether half of the contexts replace the original goal with an alternative goal while holding the state and history fixed (Table~\ref{tab:fixed-budget}). Across paired training seeds, the 95\% confidence interval of the unseen-success difference between \evoke{} and Source replay is [1.3, 9.5].

\paragraph{Competitors in the ranking.}
With the data, positives, and objective of \evoke{} fixed, the competitors ranked below the positives are replaced by the positives of the other goals posed at the same state (filled with random actions), by random valid actions, or by the valid actions least likely under $\pi_{\mathrm{boot}}$ (Table~\ref{tab:competitors}).

\begin{appendixtable}
\centering\small
\begin{minipage}[t]{.44\linewidth}
\centering
\caption{\textbf{Alternative goals under a fixed budget.} Micro success (\%).}
\label{tab:fixed-budget}
\begin{tabular}{lrr}
\toprule
\textbf{Goals} & \textbf{Seen} & \textbf{Unseen} \\
\midrule
Original goals only & 84.6 & 78.8 \\
Half alternative goals & \textbf{86.1} & \textbf{84.5} \\
\bottomrule
\end{tabular}
\end{minipage}\hfill
\begin{minipage}[t]{.54\linewidth}
\centering
\caption{\textbf{Competitors in the ranking.} Micro success (\%) and average unseen steps.}
\label{tab:competitors}
\setlength{\tabcolsep}{3.7pt}
\begin{tabular}{lrrr}
\toprule
\textbf{Ranked against} & \textbf{Seen} & \textbf{Unseen} & \textbf{Steps} \\
\midrule
Policy-favored actions (\evoke{}) & \cellcolor{evokebest}\textbf{92.1} & \cellcolor{evokebest}\textbf{91.8} & \cellcolor{evokebest}\textbf{11.4} \\
Other goals' positives & \cellcolor{evokesecond}\underline{85.7} & \cellcolor{evokesecond}\underline{79.9} & \cellcolor{evokesecond}\underline{16.7} \\
Random actions & 78.6 & 74.6 & 18.2 \\
Least likely actions & 72.1 & 67.2 & 21.8 \\
\bottomrule
\end{tabular}
\end{minipage}
\end{appendixtable}

\begin{appendixtable}
\centering\small
\caption{\textbf{Examples of goal-directed execution} from paired unseen trajectories; parentheses give action indices.}
\label{tab:qualitative-decisions}
\begin{tabular}{@{}>{\raggedright\arraybackslash}p{.16\linewidth}>{\raggedright\arraybackslash}p{.42\linewidth}>{\raggedright\arraybackslash}p{.35\linewidth}@{}}
\toprule
\textbf{Goal} & \textbf{EVOKE} & \textbf{Positive SFT}\\
\midrule
Clean a bowl; place in cabinet & Takes the bowl (3), cleans it at the sink (5), and places it in the cabinet (7). & Repeatedly visits countertops; reaches the 50-action limit.\\\addlinespace[4pt]
Put a hot cup in cabinet & Takes the cup out of the cabinet (6), heats it in the microwave (8), and returns it to the cabinet (10). & Revisits a countertop and repeats ineffective moves; reaches the 50-action limit.\\\addlinespace[4pt]
Put two keychains in safe & Places the first keychain (5), finds and takes a second keychain (14), and places it in the same safe (16). & Shares actions 1--5, then revisits locations and picks up a watch (21); reaches the 50-action limit.\\
\bottomrule
\end{tabular}
\end{appendixtable}

\Needspace{150pt}
\section{Diagnostic Details}
\subsection{Consequence Probes}
\label{app:probe}
Probe data come from 120 training games (probe training), 36 further training games (layer and regularization selection), and all 134 unseen games (test), with task types stratified and no game shared across splits. In each game, we take up to eight states along a trajectory and, at each state, up to six take, put, or move actions, including actions that fail, and execute them in the simulator to obtain the post-action labels. Besides the trained policies, we probe the original Qwen2.5-3B-Instruct without any adapter. The input contains the goal, the history before the action, and the candidate action, without the list of available actions or any post-action information. We extract the hidden state at the last action token from layers 9, 18, 27, and 36 and fit class-balanced logistic regression with L2 strength in $\{10^{-4},10^{-3},10^{-2}\}$, selecting the layer and strength on the selection games. The action-text baseline fits the same classifier to unigram and bigram features of the action, and the pre-action baseline predicts that the label is unchanged by the action. For hot, clean, and cool, probe data come from the heat, clean, and cool task families of the same game splits, using heat, clean, and cool actions from states along reference trajectories and states next to the relevant appliance. Labels are read from the simulator after execution; the test set contains 1,110 held/contained transitions from 134 unseen games and 2,102 hot/clean/cool transitions from 75 unseen games. ``Same action'' restricts the test set to actions whose text appears with different outcomes in the same game; Figure~\ref{fig:knowledge-use}a reports this subset, and Table~\ref{tab:probe} gives both.

\begin{appendixtable}
\centering\small\setlength{\tabcolsep}{4pt}
\caption{\textbf{Consequence probes on unseen games.} Balanced accuracy (\%) on all test transitions and on the same-action subset.}
\label{tab:probe}
\begin{tabular}{lcccccccccc}
\toprule
& \multicolumn{2}{c}{\textbf{Hot}} & \multicolumn{2}{c}{\textbf{Held}} & \multicolumn{2}{c}{\textbf{Cool}} & \multicolumn{2}{c}{\textbf{Clean}} & \multicolumn{2}{c}{\textbf{Contained}} \\
\cmidrule(lr){2-3}\cmidrule(lr){4-5}\cmidrule(lr){6-7}\cmidrule(lr){8-9}\cmidrule(lr){10-11}
\textbf{Input} & \textbf{All} & \textbf{Same} & \textbf{All} & \textbf{Same} & \textbf{All} & \textbf{Same} & \textbf{All} & \textbf{Same} & \textbf{All} & \textbf{Same} \\
\midrule
Action text & 64.9 & 52.5 & 68.5 & 52.6 & 54.4 & 47.3 & 56.4 & 46.6 & 70.1 & 49.5 \\
Pre-action state & 86.9 & 85.6 & 72.7 & 77.0 & 83.4 & 82.6 & 85.1 & 85.1 & 60.4 & 66.4 \\
\midrule
Qwen2.5-3B & 99.5 & 99.6 & 99.1 & 97.7 & 98.8 & 98.6 & 99.5 & 99.6 & 99.3 & 98.4 \\
$\pi_{\mathrm{boot}}$ & 99.4 & 98.8 & 99.5 & 98.8 & 98.4 & 98.1 & 99.8 & 100.0 & 99.3 & 98.4 \\
Positive SFT & 99.7 & 99.8 & 99.1 & 98.3 & 98.4 & 98.3 & 99.8 & 100.0 & 99.1 & 98.1 \\
\evoke{} & 99.6 & 99.3 & 99.4 & 98.4 & 98.5 & 98.7 & 99.8 & 100.0 & 98.9 & 97.9 \\
\bottomrule
\end{tabular}
\end{appendixtable}

\Needspace{210pt}
\paragraph{Probe controls.} Three controls use the held and contained data and the same probe protocol (Table~\ref{tab:probe-controls}). A model with the Qwen2.5-3B architecture and random weights tests whether decodability requires pretraining. A control task \citep{hewitt2019control} assigns each action type (verb, object class, and receptacle class) a fixed random label with the base rate of the real label, so that accuracy on it reflects what the probe can memorize. A history-text baseline fits the classifier to unigram and bigram features of the goal, history, and action.

\begin{appendixtable}
\centering\small\setlength{\tabcolsep}{5pt}
\caption{\textbf{Probe controls on unseen games.} Balanced accuracy (\%).}
\label{tab:probe-controls}
\begin{tabular}{lcccc}
\toprule
& \multicolumn{2}{c}{\textbf{Held}} & \multicolumn{2}{c}{\textbf{Contained}} \\
\cmidrule(lr){2-3}\cmidrule(lr){4-5}
\textbf{Input} & \textbf{All} & \textbf{Same} & \textbf{All} & \textbf{Same} \\
\midrule
Action text & 68.5 & 52.6 & 70.1 & 49.5 \\
History text & 80.9 & 77.1 & 75.7 & 71.3 \\
Random weights & 66.1 & 65.0 & 69.2 & 68.0 \\
Qwen2.5-3B & 99.1 & 97.7 & 99.3 & 98.4 \\
Qwen2.5-3B, control task & 60.5 & 56.9 & 66.7 & 60.2 \\
\bottomrule
\end{tabular}
\end{appendixtable}

\Needspace{150pt}
\subsection{Goal-Conditioned Action Choice}
\label{app:goal-swap}
We take states at seven evenly spaced positions along reference trajectories of the 134 unseen games and pair each state with up to three alternative goals. A pair is kept if both goals are achievable and not yet satisfied, the state, history, and available actions are identical under both goals, and every available action is labeled as advancing each goal or not. This yields 986 pairs from 109 games; the 812 pairs whose two goals have disjoint positive sets form the evaluation set of Figure~\ref{fig:knowledge-use}b. Each model scores every available action by its mean token log-probability, and its top-ranked action is taken as its choice. A decision counts as a habitual error when the top-ranked action belongs to the positive set of the other goal in the pair; the rate is computed over both decisions of every pair. Relative to $\pi_{\mathrm{boot}}$, source replay, and positive SFT, \evoke{} improves joint accuracy by 27.3 [23.6, 31.2], 22.4 [18.9, 26.0], and 3.2 [0.2, 6.2] points and reduces habitual errors by 16.9 [15.0, 18.8], 6.9 [4.8, 9.0], and 4.3 [3.0, 5.6] points, respectively (95\% game-level bootstrap intervals). Choosing uniformly among the available actions solves 0.3\% of pairs. Goal pairs and task success measure different abilities. A goal pair isolates one decision, the same state, history, and available actions under two goals, whereas success on unseen games also depends on execution, such as issuing valid actions and not revisiting locations (Figure~\ref{fig:execution}). Source replay executes far better than $\pi_{\mathrm{boot}}$ (4.2 vs.\ 11.9 invalid actions per game) while leaving its per-decision accuracy on goal pairs unchanged (55.4\%, Figure~\ref{fig:knowledge-use}b); Positive SFT chooses by the goal more often but makes more invalid actions and revisits than Source replay. \evoke{} improves both.

%% file: tables/controlled_task_breakdown.tex
\begin{appendixtable}
\centering\small
\caption{\textbf{Taskwise results of the decision-supervision ablations.} Success (\%) on Seen and Unseen games; averaged over training seeds.}
\label{tab:tasks}
\begin{tabular}{lrrrrrrrr}
\toprule
\textbf{Method} & \textbf{Pick} & \textbf{Look} & \textbf{Clean} & \textbf{Heat} & \textbf{Cool} & \textbf{Pick2} & \textbf{Micro} & \textbf{Macro} \\
\midrule
\multicolumn{9}{c}{\textbf{Seen}} \\
\midrule
$\pi_{\mathrm{boot}}$ & 74.3 & 61.5 & 77.8 & 62.5 & 60.0 & 70.8 & 69.3 & 67.8 \\
Source only & 91.4 & 84.6 & 81.5 & 81.3 & 68.0 & 87.5 & 82.9 & 82.4 \\
Source replay & 95.4 & 95.9 & 90.9 & 92.5 & 73.9 & 95.3 & 90.4 & 90.6 \\
Positive SFT & 97.1 & 84.6 & 81.5 & 87.5 & 96.0 & 95.8 & 91.4 & 90.4 \\
\evoke{} & 91.4 & 76.9 & 96.3 & 100.0 & 92.0 & 91.7 & 92.1 & 91.4 \\
\midrule
\multicolumn{9}{c}{\textbf{Unseen}} \\
\midrule
$\pi_{\mathrm{boot}}$ & 50.0 & 33.3 & 64.5 & 78.3 & 71.4 & 58.8 & 60.4 & 59.4 \\
Source only & 91.7 & 88.9 & 74.2 & 87.0 & 71.4 & 88.2 & 82.8 & 83.6 \\
Source replay & 88.9 & 85.2 & 88.8 & 86.1 & 80.0 & 83.9 & 85.9 & 85.5 \\
Positive SFT & 79.2 & 77.8 & 83.9 & 87.0 & 95.2 & 82.4 & 84.3 & 84.2 \\
\evoke{} & 95.8 & 72.2 & 93.6 & 95.7 & 95.2 & 94.1 & 91.8 & 91.1 \\
\bottomrule\end{tabular}\end{appendixtable}